\documentclass[10pt,twocolumn,letterpaper]{article}

\usepackage{cvpr}              

\usepackage{graphicx}
\usepackage{amsmath}
\usepackage{amssymb}
\usepackage{booktabs}
\usepackage[accsupp]{axessibility}

\usepackage[pagebackref,breaklinks,colorlinks]{hyperref}

\usepackage[capitalize]{cleveref}
\crefname{section}{Sec.}{Secs.}
\Crefname{section}{Section}{Sections}
\Crefname{table}{Table}{Tables}
\crefname{table}{Tab.}{Tabs.}

\def\cvprPaperID{3020} 
\def\confName{CVPR}
\def\confYear{2023}

\begin{document}

 \title{Efficient Multimodal Fusion via Interactive Prompting}

\author{Yaowei Li \textsuperscript{1}\\
\and
Ruijie Quan \textsuperscript{2}\\
\and
Linchao Zhu \textsuperscript{2}\\
\and
Yi Yang \textsuperscript{2}
\and
yaowei.li@uts.edu.au, \{quanruijie, zhulinchao, yangyics\}@zju.edu.cn
\and
\textsuperscript{1} ReLER, AAII, University of Technology Sydney\\
\textsuperscript{2} CCAI, Zhejiang University
}
%

\maketitle
\begin{abstract}
Large-scale pre-training has brought unimodal fields such as computer vision and natural language processing to a new era. 
Following this trend, the size of multimodal learning models constantly increases, leading to an urgent need to reduce the massive computational cost of finetuning these models for downstream tasks.  
In this paper, we propose an efficient and flexible multimodal fusion method, namely PMF, tailored for fusing unimodally pretrained transformers.
Specifically, we first present a modular multimodal fusion framework that exhibits high flexibility and facilitates mutual interactions among different modalities. 
In addition, we disentangle vanilla prompts into three types in order to learn different optimizing objectives for multimodal learning. 
It is also worth noting that we propose to add prompt vectors only on the deep layers of the unimodal transformers, thus significantly reducing the training memory usage.
Experiment results show that our proposed method achieves comparable performance to several other multimodal finetuning methods with less than 3\% trainable parameters and up to 66\% saving of training memory usage.
\end{abstract}
\section{Introduction}
\label{sec:intro}

\begin{figure}
  \centering
  \includegraphics[width=1.0\linewidth]{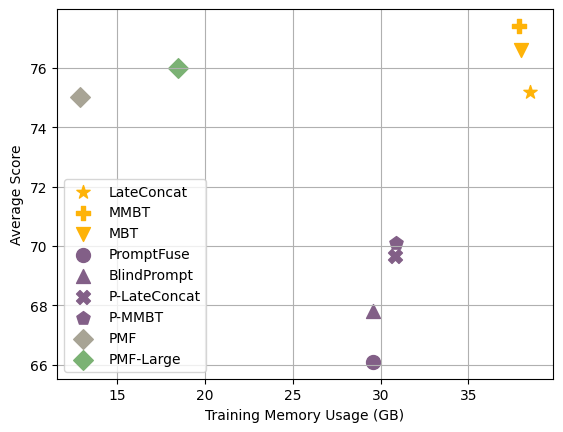}
  \caption{\textbf{Comparison over three multimodal classification tasks.} We compare our proposed PMF and PMF-Large with multiple finetuning (yellow) and prompt-based (purple) methods. The y-axis is the average score of three tasks, and the x-axis is the maximum GPU memory usage during training. 
  }
  \label{fig:Perf_Memory_scatter}
\end{figure}

Recent years have witnessed the great success of large-scale pretrained language models~\cite{radford2018improving,devlin2018bert,raffel2020exploring} and visual models~\cite{liu2021swin,chen2021exploring,he2022masked,10.1007/978-3-031-20056-4_35}, leading to a surge of pretrained multimodal models~\cite{ju2021prompting,khattak2022maple,yang2022prompt,zhou2022learning,zhu2020actbert} trying to align different modalities.
Many prior methods utilize finetuning to update the entire set of model parameters for every target cross-modal task.
Although finetuning can achieve good performance, it requires a large number of computational costs since the gradients and optimizer states for all parameters of multimodal models have to store.
Therefore, it encourages researchers to propose more parameter-efficient methods than finetuning for multimodal learning.

More recently, prompting tuning~\cite{liu2021gpt, lester2021power, li2021prefix, qin2021learning, liu2021p} is proposed to address this problem by freezing all parameters of a pretrained model while tuning only the continuous prompts.
Specifically, it adds trainable continuous prompts to the original token sequences of input data.
During training, only the continuous prompts are updated.
For multimodal prompt-based learning, a most recent method~\cite{liang2022modular} proposes to disentangle the functionality of the pretrained model which exhibits high flexibility. 
Although this method significantly reduces the tuned parameters (\eg, less than 0.1\% of the pretrained model), there still exists a large performance gap between it and the finetuning-based methods.
In addition, this method adopts a sequential modular structure that the pretrained image transformer model is followed by a language transformer model, which causes two main problems in cross-modal learning: a one-way path learning and a significant increase in the number of model layers.
Specifically, a one-way path learning in the multimodal model usually forces one modality to align with others, but not vice versa. In this way, cross-modal learning based on multiple different modalities is not fully explored due to the missing mutual alignments.
Since the prompts are added to the token sequences of input data and are updated in the training, they require extensive gradient calculations in the backward propagation which cost numerous memory usages. As a result, this kind of method does not reduce the memory usage during training by much (up to 20\%) though it reduces the number of parameters to update. In other words, this parameter-efficient method still requires massive computational resources which prevents it from being applied to many real-world applications.

To address these issues, we propose a \textbf{P}rompt-based \textbf{M}ultimodal \textbf{F}usion method with a high memory efficiency, namely \textbf{PMF}. 
Firstly, we present a new form of modular multimodal fusion framework which demonstrates high flexibility and facilitates a two-way interaction among different modalities.
Specifically, we adopt a two-stream structure where the pretrained language model and image model construct the multimodal model in a parallel way. 
Therefore, tokens of different modalities can learn mutual interactions through a cross-attention-like operation. 
Such a parallel modular structure brings two benefits. 
First, unimodal pretraining can be directly utilized for multimodal learning through a parallel combination, eliminating the need for paired multimodal datasets that can be expensive to construct.
Also, the type of image or language model can be changed easily (\eg, replacing BERT with T5 for text generation tasks). 
Furthermore, incorporating extra modalities is made possible based on the parallel modular structure.

Moreover, we propose to leverage three types of interactive prompts (\ie, query prompts, query context prompts, and fusion context prompts) in order to dynamically learn different objectives for multimodal learning.
Intuitively, the query context prompt and query prompt can be seen as a pair of `questions' and `answers' with an aim of extracting necessary information for exchange between two modalities.
After being translated by a non-linear mapping `translator', the `answer' is then delivered to the other modality for better cross-modal understanding. 
Finally, the fusion context prompts then provide the context to the delivered answer to facilitate the fusion. 

Last but most importantly, PMF is a memory-efficient method that significantly reduces the memory requirements for the large pretrained model.
Considering that calculating gradients for prompts for back-propagation is memory-consuming, we propose to add prompts only on the deep layers of the utilized unimodal transformers.
Therefore, instead of passing through the entire multimodal model, the backward propagation only needs to pass through the deep few transformer layers to reach all trainable parameters, greatly reducing the training memory usage. 
We conduct extensive experiments to demonstrate the superior of it in our experiments.
As a result, PMF enables large pretrained models to be trained on the GPU with a low memory requirement.

We conduct extensive experiments on three vision-language datasets: UPMC-Food101 \cite{wang2015recipe}, MM-IMDB\cite{arevalo2017gated}, and SNLI-VE \cite{xie2019visual}.
Through comparisons with multiple finetuning and prompt tuning methods (see in \cref{fig:Perf_Memory_scatter}), we find that: 
(1) PMF is the most memory-efficient method for cross-modal learning so far, which reduces the training memory usage by up to 66\% compared with finetuning baselines, and by 55\% compared with prompt-based methods.
(2) PMF can perform comparably compared to prior fine-tuning methods with much fewer trainable parameters (less than 2.5\%) and memory usage.

Concretely, our contributions are as follows: 
(1) we present a new form of modular multimodal fusion framework which enables two-way interactions between different modalities and high flexibility of the entire model; 
(2) we disentangle vanilla prompts into three types of prompts, in order to dynamically learn different objectives for multimodal learning;
(3) our proposed method is quite memory-efficient yet is able to achieve comparable performance with existing finetuning methods for multimodal fusion.

\section{Related works}
\label{sec:Related Work}

\noindent\textbf{Multimodal Fusion}.
Multimodal fusion methods aim to simultaneously process the input of different modalities, such as audio-video \cite{nagrani2021attention}, vision-language \cite{arevalo2017gated, kiela2019supervised}, and inputs from different types of sensors \cite{xiong2002multi}, etc. In this paper, we specifically focus on the fusion of vision-language inputs, though our proposed strategy is compatible with other modality pairs as long as there are unimodally pretrained transformers for these modalities.

Our work is in line with deep learning-based multimodal fusion strategies \cite{ngiam2011multimodal, srivastava2012multimodal, kiros2014multimodal, tsimpoukelli2021multimodal, manas2022mapl, li2021DCC,yang2021multiple}. 
In this line of work, \cite{kiela2019supervised} proposed a framework whose vision encoder solely serves as a mapping tool to encode the raw images to the token space of the text encoder. 
Such an architecture is widely used in the later multimodal fusion research \cite{tsimpoukelli2021multimodal, liang2022modular, manas2022mapl}. 
Differently, \cite{nagrani2021attention} used a dual-encoder architecture with bottleneck fusion tokens to exchange information between two encoders for video-audio fusion. Our work has a similar architecture as in \cite{nagrani2021attention}. But our proposed method completely freezes the unimodal encoders and uses an interactive prompting technique for more efficient fusion.

Another important line of multimodal fusion work is through pretraining with the large-scale datasets, mostly achieved by self-supervised learning algorithms\cite{lu2019vilbert, radford2021learning, wang2022ofa, yu2022coca, alayrac2022flamingo, jia2021scaling, wang2022bevt}. 
This line of works can be roughly divided into two kinds by their architectures. 
The first kind has a dual-encoder structure where image and text are treated separately, such as CLIP \cite{radford2021learning} and ALIGN \cite{jia2021scaling}. 
The other kind simultaneously processes the vision and language inputs with cross-attention or self-attention over a longer sequence from two modalities. 
Our proposed method differs from these self-supervised architectures in that the individual components used in our model are unimodally pretrained with much less data.
This difference directly results in a huge performance gap because of the lack of multimodal information in the pretraining stage. 
However, using unimodally pretrained models enables a much more flexible architecture. 
It has great potential in multimodal tasks where modality-paired large-scale pretraining data is not available, or when more advanced unimodal encoders are proposed in the future.

\begin{table}
  \centering
  \begin{tabular}{@{}lc@{}lc@{}}
    \toprule
    Task Type & Pretraining Type & \ \ \ \ \ \ \ \ Method \\
    \midrule
    Unimodal & Unimodal & P-Tuning\cite{liu2021gpt}, VPT\cite{jia2022visual} \\ 
    Unimodal & Multimodal & CoOp\cite{zhou2022learning}, MAPLE\cite{khattak2022maple}\\
    Multimodal & Multimodal & PI-VL\cite{ju2021prompting}, PTGM\cite{yang2022prompt}\\
    Multimodal & Unimodal & BlindPrompt\cite{liang2022modular}, \bf{PMF}\\
    \bottomrule
  \end{tabular}
  \caption{\textbf{Prompt-based methods categorized by model pretraining types and downstream task types.} Our proposed PMF falls into the last category which utilize unimodally pretrained transformers for multimodal tasks.}
  \label{tab:category}
\end{table}

\noindent\textbf{Prompt Tuning}. 
As shown in \cref{tab:category}, prompt-based methods can be roughly divided into four major categories in terms of the modalities of the pretrained model and the downstream tasks. Prompting techniques originally apply on unimodally pretrained transformers for unimodal natural language processing (NLP) tasks \cite{liu2021gpt, lester2021power, li2021prefix, qin2021learning, liu2021p}. Pretrained GPT-3 can be simply leveraged with handcrafted prompts, which are some manually chosen words preceding the input text \cite{brown2020language}. Then \cite{liu2021gpt} and \cite{lester2021power} proposed to change the handcrafted prompt to trainable continuous prompts and only update the prompt vectors during the training. Later on, \cite{li2021prefix} and \cite{qin2021learning} proposed to use prompt tuning in every hidden layer in the pretrained transformer instead of the input embeddings only. VPT \cite{jia2022visual} first applied prompt tuning to the vision transformer. 

For methods that prompt multimodally pretrained models for unimodal tasks, many recent works apply prompt tuning to pretrained vision-language models ({\em i.e.} CLIP \cite{radford2021learning}) for unimodal vision tasks \cite{zhou2022learning, zhou2022conditional, khattak2022maple, lu2022prompt, bahng2022visual}. Another type of prompt-based method apply to the multimodal pretrained model for multimodal tasks \cite{ju2021prompting, yang2022prompt}. \cite{ju2021prompting} adds prompts to an encoder-decoder one-for-all multimodal transformer, achieving comparable performance with finetuning with improved robustness against adversarial attacks. The method used in \cite{ju2021prompting} is simple but effective, showing that prompting methods works well with powerful and complex multimodally pretraining models.

Our proposed fusion strategy is different from the above methods and falls into the last category where prompts are used to fuse pretrained unimodal models for multimodal tasks. Sharing the same architecture design with \cite{kiela2019supervised, tsimpoukelli2021multimodal}, \cite{liang2022modular} uses prompt to align the feature extracted from raw images to the token space of pretrained language model. They achieved comparable performance to several multimodal fusion methods in low-resource settings but underperform fine-tuning baselines by a large margin with full data. Compared with them, our proposed PMF not only is more memory-efficient but can also perform comparably with fine-tuning baselines with full data.

\begin{figure*}
  \centering
  \includegraphics[width=1.0\linewidth]{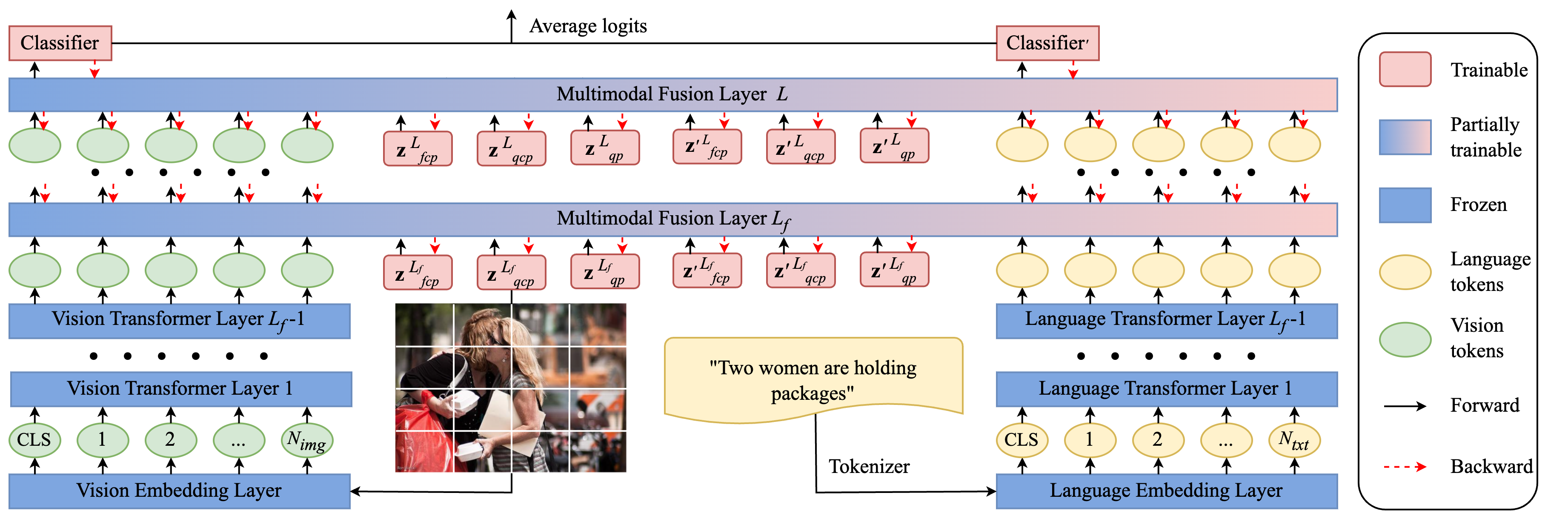}
  \caption{\textbf{Prompt-based multimodal fusion strategy (PMF) applied to vision-language inputs.} In the forward propagation, image and text inputs are first embedded into continuous token sequences and fed to the unimodal transformer layers for base feature extraction. The base features from both modalities then pass through multiple prompt-based multimodal fusion layers to get the feature of two CLS tokens for final classification. In the backward propagation, only multimodal fusion layers take part in the calculation of gradients, greatly saving memory usage during training. All pretrained parameters in both transformers are frozen during the training.}
  \label{fig:overview}
\end{figure*}

\section{Prompt-based Multimodal Fusion}
\label{sec:Method}
In this section, we describe our proposed \textbf{P}rompt-based \textbf{M}ultimodal \textbf{F}usion strategy (\textbf{PMF}). 
We begin by summarising unimodal transformers developed for vision and language tasks in \cref{sec:Unimodal Transformers}. 
Then we describe the base feature extraction process in \cref{sec:Base feature Extraction}. 
Lastly, we give a detailed description of how PMF integrates two unimodal transformer layers into a multimodal one via interactive prompting in \cref{sec:Multimodal Fusion Layer}.

\subsection{Unimodal Transformers}
\label{sec:Unimodal Transformers}

Vision Transformer (ViT) \cite{dosovitskiy2020image} adapts the Transformer \cite{vaswani2017attention} architecture with minimum modifications. 
The RGB image input $\boldsymbol{\rm{x}}_{img} \in \mathbb{R}^{h,w,c}$ is first cut into $N_{img}$ non-overlapping patches and then linearly projected into a sequence of embeddings $\boldsymbol{\rm{z}}$ with each $z_i \in \mathbb{R}^d$. 
Differently, the language Transformer first tokenizes raw text to $N_{txt}$ one-hot word embeddings and then converted these discrete vectors into a sequence of $N_{txt}$ continuous embeddings. The resulting continuous embedding for both Language Transformer and Vision Transformer share the same structure as follows:
\begin{equation}
  \boldsymbol{\rm{z}} = [{\rm CLS}, z_1, z_2, ..., z_N]
  \label{eq:input structure}
\end{equation}
where CLS is a special token prepended to the sequence so that its representation at the final layer can be used as the representation of the whole sequence for classification. Please note that the two unimodal transformers have different CLS tokens.
The continuous embedding $\boldsymbol{\rm{z}}$ is then fed into a transformer encoder which consists of $L$ transformer layers. For each transformer layer, the input passes through modules including multi-head self-attention, layer normalization, multilayer perceptron, and finally added to the original input with a residual connection. 

\subsection{Unimodal Base feature Extraction}
\label{sec:Base feature Extraction}
As shown in \cref{fig:overview}, the image and text inputs are first processed and fed into the unimodal transformer layers to extract base features, respectively. At this stage, each encoder works exactly the same as they did in unimodal tasks. Here we denote the starting fusion layer as $L_f$. And the base feature extraction of each encoder can be denoted as:
\begin{equation}
  \boldsymbol{\rm{z}}^{l+1} = {\rm{TransLayer}}^{l}(\boldsymbol{\rm{z}}^{l}; \theta) \quad \text{if $l < L_f$}
  \label{eq:feature extraction}
\end{equation}
where $\theta$ stands for the pretrained parameters. A smaller $L_f$ leads to an earlier fusion and a larger $L_f$ leads to a later fusion. A detailed discussion of the impact brought by different $L_f$ can be found in \cref{sec:ablation}.

\begin{figure*}
  \centering
  \includegraphics[width=1.0\linewidth]{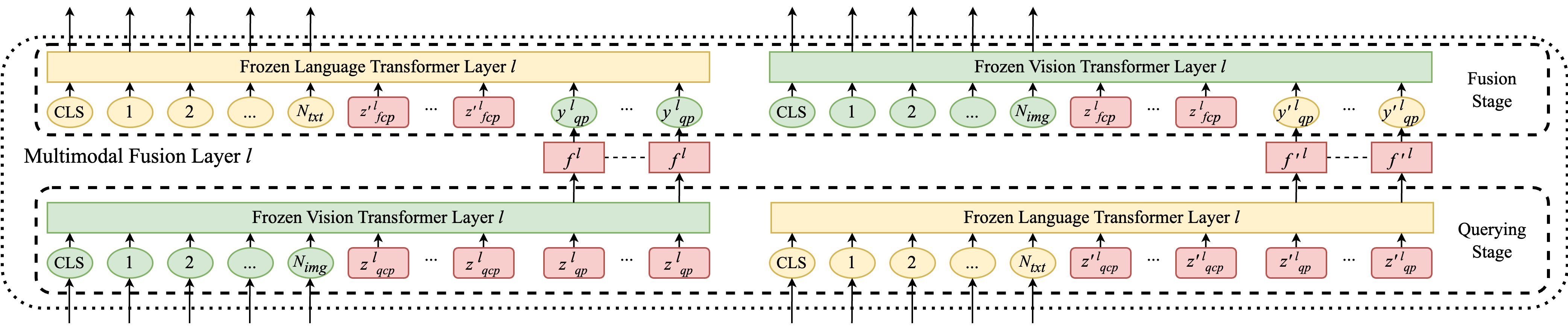}
  \caption{\textbf{Prompt-based multimodal fusion layer.} We propose to use three kinds of interactive prompts to achieve the fusion/exchange of information from two modalities. We use `query prompt' (${\rm{z}}_{qp}$, ${\rm{z}}_{qp}^\prime$) and `query context prompt' (${\rm{z}}_{qcp}$, ${\rm{z}}_{qcp}^\prime$) in the querying stage to extract what is necessary, after a non-linear mapping, the extracted information is then fused to the other modality with the help of `fusion context prompt'(${\rm{z}}_{fcp}$, ${\rm{z}}_{fcp}^\prime$) in the fusion stage. Yellow and green elements stand for language and vision modalities, respectively. Red boxes indicate trainable modules.}
  \label{fig:fusion_layer}
\end{figure*}

\subsection{Multimodal Fusion Layer}
\label{sec:Multimodal Fusion Layer}
The extracted unimodal base features are then passed through multiple multimodal fusion layers, each consisting of a `querying stage' and a `fusion stage', as shown in \cref{fig:fusion_layer}
The querying stage focus on the extraction of what is necessary to pass, and the fusion stage focus on fusing the extracted information delivered from the other modality.

This two-stage setting makes the vanilla prompt training entangled with different learning objectives. 
Therefore, we propose to decouple the vanilla prompts into three kinds: `query prompt' (QP, denoted $\boldsymbol{\rm{z}}_{qp}$), `query context prompt' (QCP, denoted $\boldsymbol{\rm{z}}_{qcp}$), and `fusion context prompt' (FCP, denoted $\boldsymbol{\rm{z}}_{fcp}$) to dynamically learn different objectives for multimodal learning.
According to the modality where prompts are used, each kind of prompt can be further specified as $\boldsymbol{\rm{z}}_{*}$ and $\boldsymbol{\rm{z}}^{\prime}_{*}$ to distinguish from each other (\eg $\boldsymbol{\rm{z}}_{fcp}$ and $\boldsymbol{\rm{z}}^{\prime}_{fcp}$).

As shown in \cref{fig:fusion_layer}, QP and QCP are used in the querying stage and the FCP is used in the fusion stage. 
As suggested by their names, QP is to query information from the unimodal input sequence, QCP is to help this process by providing extra context to the query. QP and QCP like a pair of `questions' and `answers', translated by the non-linear mapping. As for FCP, it is responsible for providing the context to the fusion in the fusion stage. We now introduce how these three kinds of prompts interact with each other in the two stages of each multimodal fusion layer.

\noindent\textbf{Querying Stage.} 
We first concatenate corresponding QP and QCP to the input sequence $\boldsymbol{\rm{z}}$. The resulting input sequence after the concatenation is:
\begin{equation}
[\boldsymbol{\rm{z}}^{l} || \boldsymbol{\rm{z}}_{qcp}^{l} || \boldsymbol{\rm{z}}_{qp}^{l}]
\label{eq:concat input}
\end{equation}
where '$||$' denotes the concatenation operation. 
Then we feed the concatenated sequence to the unimodal transformer layer, which can be denoted as:
\begin{equation}
  [\hat{\boldsymbol{\rm{z}}}^{l} || \hat{\boldsymbol{\rm{z}}}_{qcp}^{l} || \hat{\boldsymbol{\rm{z}}}_{qp}^{l}] = {\rm TransLayer}^l([{\boldsymbol{\rm z}}^{l} || \boldsymbol{\rm z}_{qcp}^{l} || \boldsymbol{\rm z}_{qp}^{l}]; \theta)
  \label{eq:First forward}
\end{equation}

After the forward propagation, the output of QP $\hat{\boldsymbol{\rm{z}}}_{qp}^l$ is extracted as the queried information to be used in the following fusion operations. 
It should be noted that though $\hat{\boldsymbol{\rm{z}}}_{qcp}^l$ will not be used in the following fusion operations, it has played an important role by providing the context of query in the querying stage. 

The queried fusion intermediate $\hat{\boldsymbol{\rm{z}}}_{qp}^l$ is then mapped to the representation space of the other modality through a non-linear mapping function:
\begin{equation}
  \boldsymbol{\rm{y}}_{qp}^l  = f^l(\hat{\boldsymbol{\rm{z}}}_{qp}^l)
  \label{eq:mapping}
\end{equation}
where $f$ is a non-linear mapping function. Specifically, a mapping function consists of two linear layers with a bottleneck structure to reduce dimension and only the first linear layer has a ReLu function. 
Each fusion layer $l$ has two mapping functions $f^l$ and $f^{\prime l}$, building a two-way interaction among different modalities. Note that the non-linear mapping functions contain more than 95\% of the trainable parameters in PMF.

\noindent\textbf{Fusion Stage.} 
We first concatenate the mapped fusion intermediates $\boldsymbol{\rm{y}}_{qp}^l$ to the original input sequence $\boldsymbol{\rm{z}}^{\prime l}$ and FCP $\boldsymbol{\rm{z}}^{\prime l}_{fcp}$ of the other modality.
Then we feed the concatenated sequence to the unimodal transformer layer of the other modality to complete a one-way fusion. These two processes can be together denoted as:
\begin{equation}
  [\boldsymbol{\rm z}^{\prime l+1} || \hat{\boldsymbol{\rm{z}}}^{\prime l}_{fcp} || \hat{\boldsymbol{\rm{y}}}_{qp}^l] = {\rm TransLayer}^l([\boldsymbol{\rm z}^{\prime l} || \boldsymbol{\rm z}^{\prime l}_{fcp} || \boldsymbol{\rm y}_{qp}^l]; \theta^\prime)
  \label{eq:Second forward}
\end{equation}
where $[\boldsymbol{\rm{*}}]^\prime$ means $[\boldsymbol{\rm{*}}]$ from the other modality. 

Finally, the output of two unimodal transformer layers $\boldsymbol{\rm{z}}^{l+1}$ and $\boldsymbol{\rm{z}}^{\prime l+1}$ are together as the output of the multimodal fusion layer and fed to the higher layers. The entire multimodal fusion process can be concluded as:
\begin{equation}
  [\boldsymbol{\rm z}^{l+1}||\boldsymbol{{\rm z}}^{\prime l+1}] = {\rm Fusion Layer}^l([\boldsymbol{\rm z}^{l}||\boldsymbol{\rm z}^{\prime l}]; \theta, \theta^\prime) \quad \text{if $L_f\leq l$}
  \label{eq:multimodal fusion}
\end{equation}

After the multimodal fusion is complete, we take the output representation of CLS token $z^L_{CLS}$ and $z^{\prime L}_{CLS}$ to two different linear classifiers and average the pre-softmax logits for classification. Mathematically, such a classifier setting is equivalent to feeding the classifier with the concatenated features when a linear classifier is used, except for a different scale of gradient caused by the averaging operation.

\begin{table*}[]
    \centering
    \begin{tabular}{llllll|l}
    \toprule
    Method      & \multicolumn{1}{c}{\begin{tabular}[c]{@{}c@{}}Updated Param. \\ (Million)\end{tabular}} & \multicolumn{1}{c}{\begin{tabular}[c]{@{}c@{}}Memory Usage (GB) \\ \midrule Train/Inference \end{tabular}} & SNLI-VE & Food-101 & MM-IMDB & Avg.\\
    \midrule
    Linear      &    -                  & 3.76 / 3.23    &    50.05    &     78.96    &     49.76 / 56.83 &  60.77  \\ \hline
    ViT         &    86.5               & 9.36 / 1.99    &    33.33    &     74.69    &     38.39 / 49.88 &  50.72 \\
    BERT        &    109.0              & 30.82 / 2.79   &    69.82    &     87.44    &     58.91 / 64.31 &  72.96 \\
    LateConcat  &   196.0               & 38.54 / 3.36   &    70.01    &     93.29    &     59.56 / 64.92 &  75.18\\
    MMBT$^*$    &   196.5               & 37.87 / 3.48   &    74.69    &     94.10    &     60.80 / 66.10 &  77.41 \\
    MBT$^*$     &   196.0               & 38.00 / 4.06   &    74.02    &     93.56    &     59.60 / 64.81 &  76.60  \\ \hline
    VPT         &    -                  & 6.12  / 2.01   &    33.33    &     72.55    &     35.22 / 44.49 &  48.58 \\
    P-BERT      &    -                  & 28.13 / 2.99   &    63.28    &     81.07    &     48.67 / 54.58 &  65.33  \\
    PromptFuse  &    -                  & 29.57 / 3.55   &    64.53    &     82.21    &     48.59 / 54.49 &  66.09 \\
    BlindPrompt &    -                  & 29.57 / 3.65   &    65.54    &     84.56    &     50.18 / 56.46 &  67.81 \\
    P-LateConcat&    0.3                & 30.82 / 3.43   &    63.05    &     89.03    &     53.91 / 59.93 &  69.67   \\
    P-MMBT      &    0.9                & 30.90 / 3.48   &    67.58    &     86.58    &     52.95 / 59.30 &  70.10 \\ \hline
    \textbf{PMF} ($M$=4, $L_f$=10) &2.5 &12.84 / 4.08 &    71.92    &     91.51    &     58.77 / 64.51 &  75.02   \\
    \textbf{PMF-large} ($M$=4, $L_f$=22) &4.5 &18.44 / 6.42 & 72.10 &    91.68    &     61.66 / 66.72 &  75.99\\
    
    \bottomrule
    \end{tabular}
    \caption{\textbf{Multimodal classification performance.} PMF achieve comparable performance to the finetuning baselines with less than 3\% of trainable parameters and up to 66\% of training memory usage. MM-IMDB is F1-Macro / F1-Micro, others are accuracy. We report the maximum memory usage in training and evaluating UPMC Food-101 for each method. We report mean performance over 3 runs with different random seeds. `-' means trainable parameter less than 0.1 M. PMF-Large uses bert-large and vit-large models (24 hidden layers) while others use bert-base and vit-base models (12 hidden layers). $M$ is the prompt length and $L_f$ is the starting fusion layer.}
\label{tab:Main Table}
\end{table*}

\section{Experiments}
In this section, we analyze the performance of our proposed PMF on three multimodal datasets and aim to answer the following questions: (1) How efficient is the PMF and does PMF perform well (\cref{sec:main results})? (2) What factors affect the effectiveness of PMF (\cref{sec:ablation})?
In addition, we explore the PMF equipped with larger transformers and its impact brought to the performance and memory efficiency in \cref{sec:Modularity}. 
Lastly, we introduce a Neural Architecture Search (NAS) method to automatically search for the preferable fusion structure for PMF in \cref{sec:nas}.

\subsection{Datasets and Metrics}
\noindent\textbf{UPMC Food-101}\cite{wang2015recipe} is a multimodal classification dataset, which contains food images with textual recipe descriptions for 101 kinds of food. 
UPMC Food-101 contains a total of 90,840 image-text pairs with a size range between 790 and 956 pairs for different classes. As the dataset only has training and testing sets, we follow \cite{kiela2019supervised} and create a validation split of 5000 samples from the training set. 
 
\noindent\textbf{MM-IMDB}\cite{arevalo2017gated} is a multimodal multi-label classification dataset, which contains movie plot outlines and movie posters. The task is to predict the genre of movies. This dataset contains 25,956 image-text pairs for 23 classes with a long-tail distribution. 

\noindent\textbf{SNLI-VE}\cite{xie2019visual} is a multimodal classification dataset for the visual entailment tasks, which is to reason about the relationship between an image premise and a text hypothesis into entailment, contradiction, or neutrality. SNLI-VE contains a total of 565,286 image-text pairs. Please note that we only use image premise and text hypothesis in the input, which is different from the settings in some other papers where text premises are also used in the inputs \cite{yang2022prompt}. 

We report accuracy for UPMC Food-101 and SNLI-VE, and Macro/Micro-F1 scores for MM-IMDB as metrics.

\subsection{Existing Methods and Baselines}
\label{sec:baselines}
We report the performance of several baselines and existing methods. First, we report the performance of finetuning unimodal models ({\em i.e.} \textbf{BERT} \cite{devlin2018bert}, \textbf{ViT}\cite{dosovitskiy2020image}) to verify the effectiveness of multimodal fusion. 
Specifically, we take the output representation of CLS token of the last layer in ViT and BERT, and feed it into a linear classifier. 
We also report the performance of \textbf{VPT}\cite{jia2022visual} and a prompt-based BERT (denoted \textbf{P-BERT}) for a better comparison. 
For VPT and P-BERT, the input sequence to each transformer layer is concatenated with a prompt vector, whose length is set to 10. And the concatenated prompt vectors and the final linear classifier are the only updated modules in training.

In addition, we compare against a strong baseline method which concatenates the output features of CLS tokens of ViT and BERT, and feed the concatenated feature to a linear classifier, denoted as \textbf{LateConcat}. In this case, the input to the classifier is (768 + 768)-dimensional. Besides, we also introduce \textbf{Linear}, which shares the same architecture with LateConcat with the only difference in the updated modules. Linear only updates the linear classifier while LateConcat updates all parameters during training.

We reimplement MMBT (denoted \textbf{MMBT$^*$}) \cite{kiela2019supervised} and MBT (denoted \textbf{MBT$^*$})\cite{nagrani2021attention} with a \verb|vit-base| model as the vision encoder and a \verb|bert-base| model as the text encoder for fair and controlled comparison. We set the fusion layer $L_f = 8$ and use 4 fusion tokens in MBT as recommended in the original paper. 

We also propose a prompt-based MMBT and a prompt-based LateConcat, denoted as \textbf{P-MMBT} and \textbf{P-LateConcat}, respectively. In P-MMBT and P-LateConcat, we apply deep prompt tuning on both vision and language encoders, which are pretrained backbones with frozen parameters during training. We set the prompt length in each layer of two encoders to 10, totalling 240 prompt vectors. Similar to VPT and P-BERT, P-LateConcat only updates the final linear classifier and prompt vectors during training. Compared with P-LateConcat, P-MMBT have an extra linear projection layer and a smaller linear classifier to train.

Lastly, we report the performance of \textbf{PromptFuse} and \textbf{BlindPrompt} \cite{liang2022modular}, both proposed in the only existing paper which leverages unimodally pretrained models for multimodal fusion through prompting. We set the prompt length to 20 as recommended in the original paper.

\subsection{Implementation Details}
\noindent\textbf{Pretrained Backbone and Initialization}. Unless otherwise noted, we use an ImageNet-21k \cite{deng2009imagenet} pretrained \verb|vit-base| model for the vision encoder and a \verb|bert-base-uncased| model for the language encoder in all experiments. All pretrained checkpoints are from huggingface \cite{wolf2020transformers}. All prompt vectors are initialized through a Gaussian distribution (mean=0, std=0.02).

\noindent\textbf{Network Training}. We use SGD optimizer in all experiments with momentum set to 0.9 and weight decay set to $1e^{-4}$. The batch size is set to 64 for SNLI-VE, and 32 for UPMC Food-101 and MM-IMDB. Cross entropy loss is applied in all experiments and the class labels are weighted by their inverse frequency for UPMC Food-101 and MM-IMDB. More details are in the supplementary material.

\subsection{Main Results}
\label{sec:main results}
\noindent\textbf{PMF is most memory-efficient.}
As shown in \cref{tab:Main Table}, PMF is the most memory-efficient multimodal fusion model of all existing prompt-based methods and baselines. PMF can save up to 66\% of training memory usage compared with finetuning baselines. Even compared with the existing most memory-efficient prompt-based multimodal method, PMF still saves an extra of more than 50\% of training memory. 

\noindent\textbf{PMF outperforms all existing prompt-based methods.}
With the same pretrained unimodal transformers, prompt-based methods underperform the full finetuning methods by a large margin. Some prompt-based methods even underperform unimodal finetuning baselines. Our proposed PMF achieves the best performance among all prompt-based methods. Especially, PMF outperforms all unimodal baselines in all experiments, showing that the two modalities are successfully fused.

\noindent\textbf{PMF is competitive with finetuning baselines.}
\cref{tab:Main Table} shows that PMF achieves comparable performance with full finetuning baselines with less than 3\% trainable parameters while saving 66\% of memory cost, significantly narrowing the gap between finetuning and prompt-based methods. 
Furthermore, PMF even outperforms the finetuning LateConcat when equipped with larger transformers (\ie \verb|bert-large| and \verb|vit-large|). 

\begin{table}[]
\centering

\begin{tabular}{lllll}
\toprule
QP & Mapping \textit{f} &QCP & FCP &  MM-IMDB \\
\midrule
   &     &   &                                          & 48.80/56.38       \\ \hline
   &     &   &          \checkmark                      & 45.38/53.34      \\
\checkmark   &     &     &                              & 43.56/51.43       \\
\checkmark   &  \checkmark    &     &                   & 54.92/62.25        \\
\checkmark\checkmark   & \checkmark         &    &      & 57.98/63.69        \\
\checkmark   &  \checkmark    & \checkmark   &          & 58.30/64.07      \\
\checkmark     & \checkmark      &     &  \checkmark    & 58.34/64.15        \\
\checkmark   & \checkmark &  \checkmark  &  \checkmark  & 58.63/64.23       \\
\bottomrule

\end{tabular}
\caption{\textbf{PMF Component Ablation.} There are four types of trainable modules in our proposed PMF. We set the fusion layer $L_f=10$ and add different components one at a time to see their individual impact. All prompts with \checkmark have a length of 4, and prompts with \checkmark\checkmark have a length of 8.}
\label{tab:Module Ablation}
\end{table}

\begin{figure}
  \centering
  \includegraphics[width=1.0\linewidth]{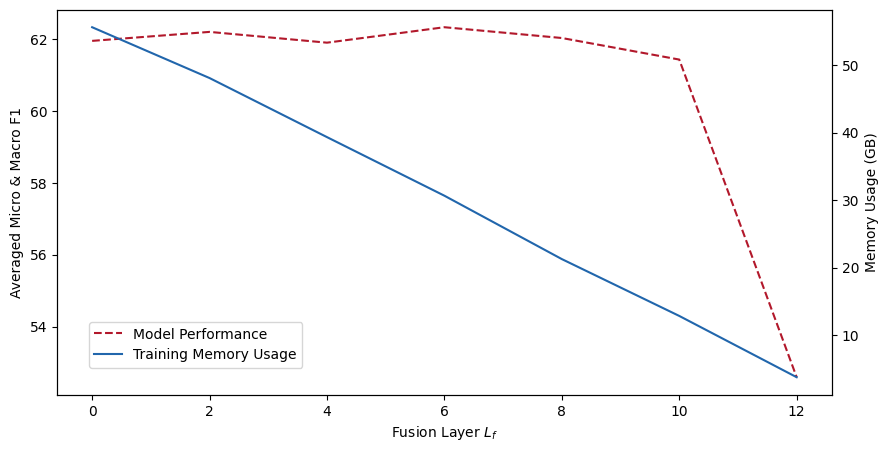}
  \caption{\textbf{Model Performance and Training Memory Usage under different fusion layers $L_f$.} We set the prompt length $M=4$ for every prompt vector with different fusion layer $L_f=0,2,4,6,8,10,12$.}
  \label{fig:layer_ablation}
\end{figure}

\begin{figure}
  \centering
  \includegraphics[width=1.0\linewidth]{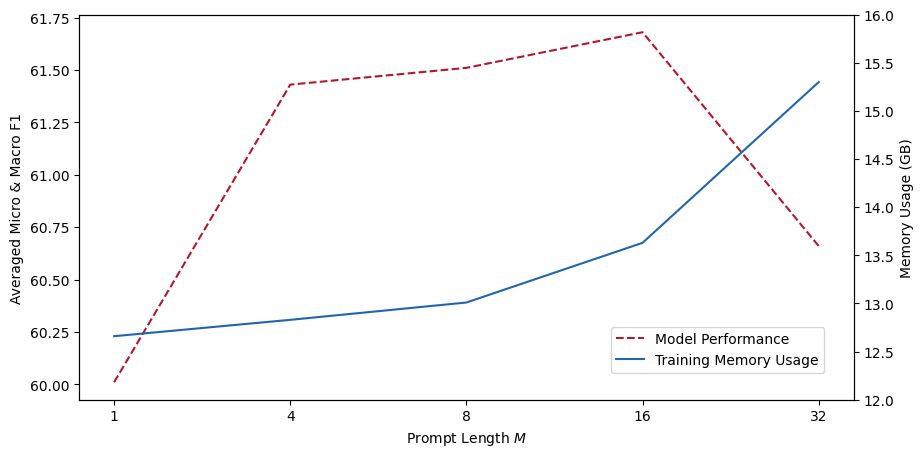}
  \caption{\textbf{Model Performance and Training Memory Usage under different prompt length $M$.} We fix the fusion layer $L_f=10$ and study 1, 4, 8, 16, and 32 tokens for each prompt vector.}
  \label{fig:length_ablation}
\end{figure}

\subsection{Ablation Study}
\label{sec:ablation}
In this section, we investigate the impact of different factors in our proposed fusion strategy. All experiments in this section are conducted on the MM-IMDB dataset.

\noindent\textbf{Components Ablation.} 
We verify the effectiveness of three kinds of prompts and the non-linear mapping function in this section.
The results are shown in \cref{tab:Module Ablation}.
The first row without any components in PMF is equivalent to the Linear model introduced in \cref{sec:baselines}. 
The comparison between the first three rows shows that merely prompting the top layers of the two transformers not only cannot achieve multimodal fusion. 
Oppositely, it will disturb the feature space of two transformers, which finally hurts the performance.
Though the mapping functions give the biggest boost to the performance, it should be noted that the mapping functions $f$ cannot work without QP querying the fusion intermediates.

In addition, the comparison between the last four rows shows that decoupling the prompts into three individual modules with different learning objectives brings a performance gain. 
More specifically, the comparison between the fifth and sixth row in \cref{tab:Module Ablation} shows that an extended QP cannot replace QCP.
Since only the output of QP tokens are fused to the other modality while that of QCP discarded, replacing QCP with a longer QP not only increases the computation as the sequences in the fusion stage are longer but also results in a damaged performance
As a result, every module introduced in PMF contributes to the quality of multimodal fusion. The missing of any of the four modules will bring a performance drop at different scales.

\noindent\textbf{Fusion Layer.} 
We now investigate impacts brought by different fusion layers $L_f$ to the fusion performance and memory efficiency.
The results are summarized in \cref{fig:layer_ablation}. 
As can be seen in the figure, the training memory usage keeps decreasing as the fusion starts later. 
However, the performance of the fusion model is relatively consistent with $L_f \leq 10$.
Therefore, adding prompts only on the deep layers ($10<l<L$) is empirically better for the trade-off between performance and memory efficiency.

\noindent\textbf{Prompt Length.}
The ablation study on prompt length is carried out with three kinds of prompts set to have the same length ({\em i.e.} $M_{qp} = M_{qcp} = M_{fcp}$).
\cref{fig:length_ablation} summarizes the results. 
The performance increases as the prompt length grow longer when $M \leq 16$, and drops when the prompts are too long ($M=32$).
It should be stressed that the training memory usage only increases around 1 GB as the prompt length grows from 1 to 16, which means the fusion layer $L_f$ is the major factor of the training memory usage instead of prompt length.
 
\begin{table}[]
\centering
    \begin{tabular}{cccc}
    \toprule
    \begin{tabular}[c]{@{}c@{}}Text\\ Encoder\end{tabular} & \begin{tabular}[c]{@{}c@{}}Image\\ Encoder\end{tabular} & \begin{tabular}[c]{@{}c@{}} Memory Usage\\ Train/Inference \end{tabular} & MM-IMDB \\
    \midrule
    bert-base       & vit-base   & 12.84 / 4.08         & 58.77 / 64.51 \\
    bert-base       & vit-large  & 14.16 / 4.89         & 59.70 / 65.20 \\
    bert-large      & vit-base   & 17.17 / 5.53         & 60.08 / 65.41 \\
    bert-large      & vit-large  & 18.44 / 6.42         & 61.66 / 66.72 \\
    \bottomrule
    \end{tabular}
\caption{\textbf{Comparison of PMF applying to different unimodal transformers.} We set the fusion layer $L_f = L-2$ and prompt length $M=4$ in all experiments. MM-IMDB is reported with F1-Macro / F1-Micro.}
\label{tab:flexibility}
\end{table}

\subsection{Modularity and Flexibility}
\label{sec:Modularity}
PMF is highly modular, which means it is trivial to replace the unimodal transformers when there are better ones. In this section, we first describe how to replace the unimodal transformer and then show the benefits of such flexibility through experiments with larger pretrained models. 

Since the total transformer layers, $L_{img}$ and $L_{txt}$ of each unimodal transformer are now different, the unimodal base features of two modalities now take different layers to extract, and the number of remaining layers for fusion stays the same. 
Simply, the fusion layers $L_f$ of two transformers can be further specified as $L_{f-img} = L_{img}-2$ and $L_{f-txt} = L_{txt}-2$.
In addition, the difference between different hidden dimensions $d$ is automatically handled by the non-linear mapping functions $f$.

The results shown in \cref{tab:flexibility} clearly demonstrate that PMF can be empowered by larger unimodal transformers with a very limited increase of training memory usage. 

\begin{table}[]
\small
\centering
\begin{tabular}{lllll}
\toprule
\multicolumn{1}{c}{\begin{tabular}[c]{@{}c@{}}Training \\ Memory\end{tabular}} & SNLI-VE & Food-101 & MM-IMDB & Avg.\\
\midrule
  33.36 GB     &    72.27    &   92.1     &   59.67 / 65.57  & 75.66   \\
\bottomrule
\end{tabular}
\caption{\textbf{Performance of PMF applied with NAS.} MM-IMDB is F1-Macro / F1-Micro, others are accuracy. We only report the training memory usage.} 
\label{tab:autoformer}
\end{table}

\subsection{PMF with NAS}
\label{sec:nas}
The hyper-parameters introduced in the proposed PMF are fusion layer $L_f$ and prompt length $M$. Although PMF works well without exhausting hyper-parameter tuning, it is still preferable to have specific settings for every different task and data distribution. 
In this section, we experiment with automatic fusion structure search via AutoFormer \cite{chen2021autoformer}. 
A detailed description of the search space and evolution search can be found in the supplementary material.

\cref{tab:autoformer} shows the performance of NAS-applied PMF on three datasets. With an increase in training memory usage, PMF-NAS achieves better results than regular PMF with the same vision and language encoders, greatly reducing the workload of finding the preferable fusion structure.

\section{Limitations and Future Works}
The first limitation is that PMF's performance on three datasets is still behind finetuning baselines with the same pretrained backbones, indicating more work developing prompt-based methods to fully leverage the knowledge inside the pretrained models in the future, finally achieving equivalent or surpassing results through prompting.

The second limitation is about the hyper-parameters tuning: It is preferable to decouple prompts into three kinds by their roles in multimodal fusion. However, it also brings more work to hyper-parameter tuning if someone is expecting the best results via an optimal fusion structure.

Our future research endeavours will involve further investigation of the PMF in diverse multimodal understanding tasks such as Visual Question Answering, utilizing various model architectures.
\section{Conclusion}
We propose a new form of modular multimodal fusion framework which demonstrates high flexibility and facilitates a two-way interaction among different modalities, namely PMF.
PMF leverages three types of interactive prompts in order to dynamically learn different objectives for multimodal learning. 
By adding the prompts only on the deep layers of utilized unimodal transformers, PMF can significantly reduce the memory usage of the gradient calculation in the backward propagation.
Through extensive experiments, we demonstrate that PMF is quite memory-efficient and yet able to perform comparably with existing finetuning baselines.
\section{Acknowledgement}
\label{sec:acknowledgement}

This work was supported in part by the Australian Research Council (ARC) under Grant DP200100938.
{\small
\bibliographystyle{ieee_fullname}
\bibliography{egbib}
}

\end{document}